\documentclass[11pt,a4paper]{article}
\usepackage{times,latexsym}
\usepackage{url}
\usepackage[T1]{fontenc}

\usepackage{graphicx}
\usepackage{booktabs}
\usepackage{multirow}
\usepackage{subcaption}
\usepackage{tabularx}
\usepackage{amsmath}
\usepackage{colortbl}
\usepackage{pifont}
\newcommand{\best}{\ding{72}}
\newcommand{\good}{\ding{118}}
\usepackage{wrapfig}
\usepackage{enumitem}

\usepackage[acceptedWithA]{tacl2021v1}

\usepackage{xspace,mfirstuc,tabulary}

\newif\iftaclinstructions
\taclinstructionsfalse 
\iftaclinstructions
\renewcommand{\confidential}{}
\renewcommand{\anonsubtext}{(No author info supplied here, for consistency with
TACL-submission anonymization requirements)}
\newcommand{\instr}
\fi

\iftaclpubformat 

\else

\fi

\title{OntoURL: A Benchmark for Evaluating Large Language Models on Symbolic \underline{Onto}logical \underline{U}nderstanding, \underline{R}easoning and \underline{L}earning}

\author{
Xiao Zhang$^\diamond$ 
\and Huiyuan Lai$^\diamond$ 
\and Qianru Meng$^\dagger$ 
\and Johan Bos$^\diamond$
\\ \\
$^\diamond$University of Groningen, Netherlands \\
\texttt{\{xiao.zhang, h.lai, johan.bos\}@rug.nl}
\\ \\
$^\dagger$Leiden University, Netherlands \\
\texttt{q.r.meng@liacs.leidenuniv.nl}
}

\date{}

\begin{document}
\maketitle
\begin{abstract}
Large language models have demonstrated remarkable capabilities across a wide range of tasks, yet their ability to process structured symbolic knowledge remains underexplored. To address this gap, we propose a taxonomy of ontological capabilities and introduce \textsc{OntoURL}, the first comprehensive benchmark designed to systematically evaluate LLMs' capabilities in handling ontologies---formal and symbolic representations of domain knowledge. Based on the proposed taxonomy, \textsc{OntoURL} systematically assesses three dimensions: understanding, reasoning, and learning through 15 distinct tasks comprising 57,303 questions derived from 40 ontologies across 8 domains. Experiments with 20 open-source LLMs reveal significant performance differences across models, tasks, and domains, with current LLMs showing capabilities in understanding ontological knowledge but weaknesses in reasoning and learning tasks. Further experiments with few-shot and chain-of-thought prompting illustrate how different prompting strategies affect model performance. Additionally, a human evaluation reveals that LLMs outperform humans in understanding and reasoning tasks but fall short in most learning tasks. These findings highlight both the potential and limitations of LLMs in processing symbolic knowledge and establish \textsc{OntoURL} as a critical benchmark for advancing the integration of LLMs with formal knowledge representations.
\end{abstract}

\section{Introduction\label{sec:intro}}

Ontologies play a foundational role in encoding structured domain knowledge among the most prominent symbolic frameworks, particularly in fields such as finance, the sciences, and law. They provide a formal structure through well-defined concepts (classes), relationships (e.g., hierarchies and semantic connections), and instances (individuals) \cite{gruber1993translation, noy2001ontology, mcguinness2004owl}. In recent years, as Large language models (LLMs) have transformed numerous fields with their remarkable capabilities in various tasks~\cite{wei-et-al-2022-chain, openai2023gpt4}, the interaction between ontologies and knowledge-rich LLMs has sparked significant interest, raising research into ontology-related tasks such as leveraging LLMs for ontology creation.

However, whether LLMs can truly comprehend and manipulate structured symbolic knowledge remains a subject of intense debate \cite{tang2023large, pavlick2023symbols, yan2024large, saba2024reinterpreting}. This discussion centers on whether statistical pattern recognition can replicate the symbolic representations and logical structures traditionally managed by classical knowledge representation systems. Therefore, a critical yet underexplored question arises for ontology practitioners and researchers: to what extent can LLMs understand, utilize, and construct ontologies? While several datasets have been developed for ontology-related tasks~\citep{wu-etal-2023-plms, bombieri2024llms, qin2024comprehensive, song2025peeled, he2023language, sun2024large, jiang2025hibench, e2eol2024lo, li2024ontology}, these works typically focus on one or two isolated ontology aspects and are rarely designed specifically for evaluating LLMs. Furthermore, there is an absence of a comprehensive taxonomy that systematically categorizes the ontological capabilities required across domains and tasks, thus hindering the reliable evaluation of LLMs capabilities on ontology.

To fill this gap, we investigate three key research questions: (1) can LLMs accurately memorize the fine-grained details inherent in ontologies, including concepts, hierarchical relationships, properties, and instances? (2) Can LLMs perform robust reasoning over ontologies, such as transitive superclass inference or description logic reasoning? (3) Can LLMs autonomously construct ontologies based on their rich knowledge, such as ontology hierarchy construction? 

We first propose a taxonomy of ontological capabilities for LLMs and then introduce \textsc{OntoURL}, the first comprehensive benchmark for evaluating LLMs' abilities on ontologies. \textsc{OntoURL} consists of 57,303 questions derived from 40 ontologies, encompassing 15 tasks across 8 domains. These tasks meticulously assess LLMs' proficiency in three crucial dimensions: understanding, reasoning, and learning. By evaluating 20 open-source LLMs, we gain critical insights into their strengths and limitations on  handling structured symbolic knowledge. Our primary contributions are summarized as follows:

\begin{itemize}
    \item We present a taxonomy of ontological capabilities in three dimensions: understanding, reasoning, and learning, providing a systematic framework for analyzing LLMs' interactions with structured symbolic knowledge.
    \item We introduce \textsc{OntoURL}, a benchmark comprising 57,303 questions derived from 40 ontologies, covering 15 tasks across 8 domains. Our benchmark enables rigorous evaluation on LLMs' abilities across multiple dimensions, such as conceptual understanding, logical reasoning, structure construction and alignment.
    \item We conduct in-depth pilot studies on 20 LLMs and provide comprehensive analyses across model scales, task levels, and specific domains. Some robust prompting strategies, such as few-shot and chain-of-thought prompting are also evaluated. An additional experiment on human performance are also provided. All code and data are available (Appendix~\ref{app:availability}).
\end{itemize}

\section{Background and Related Work}

\paragraph{Ontology}

Ontologies provide formal, logic-based representations of domain knowledge, organizing concepts, properties, instances, and their interrelations in a structured, symbolic form \cite{gruber1993translation}. Unlike taxonomies or controlled vocabularies which focus primarily on hierarchical groupings, or knowledge graphs which typically emphasize instance-level assertions, ontologies encode domain semantics through explicit axioms and logical constraints. 

\begin{figure*}
    \centering
    \includegraphics[width=\linewidth]{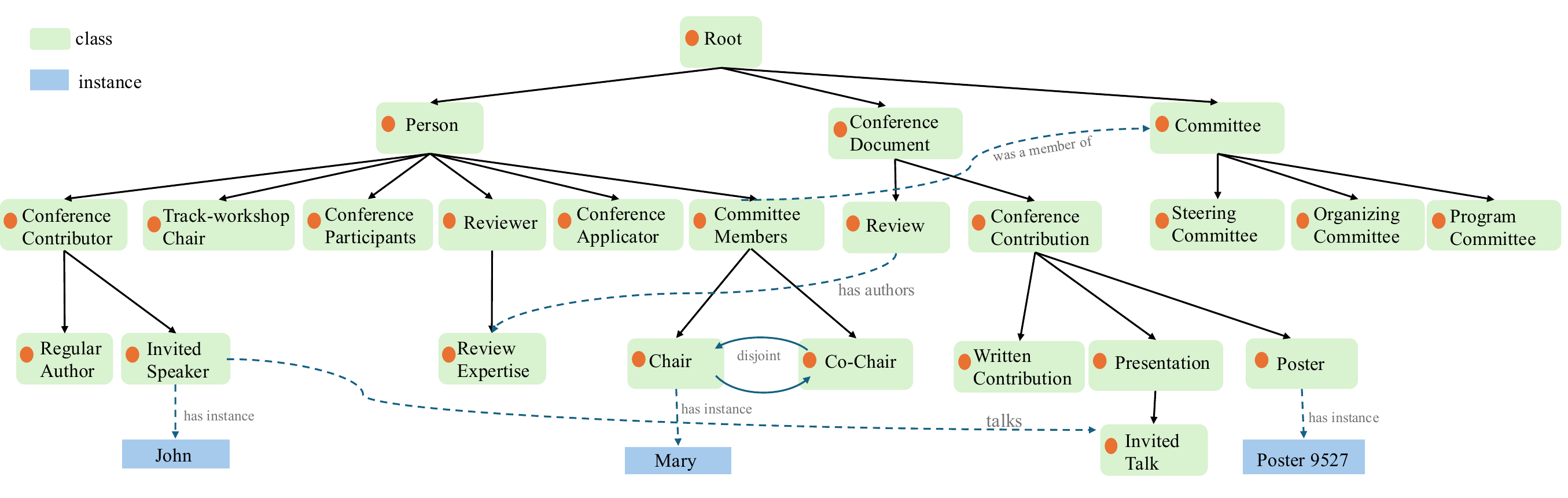}
    \caption{A sub-ontology excerpt from the Conference Ontology for representing academic conferences, illustrating the hierarchical structure of classes (in green) and instances (in blue). Most of the classes, relations, instances, and semantic information are omitted for the clarity.}
    \label{fig:ontology_example}
\end{figure*}

As a formal knowledge representation system, ontologies are typically formalized in Description Logic \cite{krötzsch2013descriptionlogicprimer}---a family of formal languages designed for representing and reasoning about knowledge. Ontologies comprise three fundamental components: Terminological Box (TBox), containing class hierarchies and definitions; Assertional Box (ABox), capturing assertions about individuals; and Role Box (RBox), defining relationships and properties among entities. Figure~\ref{fig:ontology_example} illustrates these components in a conference ontology excerpt, where  ``Chair'' and its superclass ``Committee Members'' represent TBox elements, ``Mary'' as an instance of ``Chair'' forms part of the ABox, and ``has\_authors'' is a relation between ``Review'' and ``Review Expertise'' constitutes an RBox relation.

By offering a standardized vocabulary with formal semantics, ontologies support semantic interoperability, knowledge integration, and automated reasoning \cite{staab2013handbook}. Most ontological resources are created by domain experts, such as Gene Ontology \cite{ashburner2000gene}, Plant Ontology \cite{plant2002plant}, and LKIF Core Legal Ontology \cite{hoekstra2007lkif}. 
As ontologies have been central to symbolic AI approaches for decades, understanding and leveraging such structured symbolic knowledge are essential for LLMs.

\paragraph{Ontology-related Tasks}
Previous work has primarily focused on conceptual understanding, using probing techniques to examine how LLMs memorize and retrieve class-level knowledge \cite{badie2017formal,peng-etal-2022-copen,sahu2022unpacking,patel2022mapping,wu-etal-2023-plms,shani-etal-2023-towards,jang-lukasiewicz-2023-improving,mitchell2023debate,jin2024exploring,song2025peeled}, and structural knowledge \cite{he2023language, mruthyunjaya2023rethinking,park2024geometry,jackermeier2024dual,zhang-etal-2025-neural}. 

Beyond basic understanding tasks, some works perform specialized forms of deductive logic reasoning on ontologies. Rule-based ontology reasoners support a wide range of inference tasks, including validating ontology coherence, deriving complete class hierarchies, assigning individuals to their most specific types, inferring property relationships, and executing queries to retrieve relevant classes or individuals \cite{tsarkov2006fact, mendez2009reintroducing, kazakov2012elk, sertkaya2013elephant, glimm2014hermit, ceylan2015bayesian, bobillo2016fuzzy, fernandes2018graph, balhoff2018arachne}. Recently, a few studies have begun to explore using language models for ontology reasoning, particularly within the framework of Description Logic \cite{he2023language, wang2024can}.

Another active research direction involves learning and constructing ontological structures. Traditional approaches employed statistical term extraction and pattern-based methods to identify candidate concepts and taxonomic relations \cite{hearst1998automated, kietz2000method, maedche2001ontology, xu2002domain, alfonseca2002unsupervised, lonsdale2002peppering, khan2002ontology, biemann2005ontology, asim2018survey, xu2019automatic, konys2019knowledge}. More recent efforts leverage pretrained language models, enabling more sophisticated concept extraction and hierarchy learning \cite{babaei2023llms4ol, neuhaus2023ontologies, e2eol2024lo}.

\paragraph{Ontology-related Benchmarks} 

To evaluate these diverse ontology-related tasks, researchers have developed various benchmarks focusing on specific capabilities. These include evaluations for conceptual knowledge \cite{badie2017formal, peng-etal-2022-copen, wu-etal-2023-plms, bombieri2024llms, qin2024comprehensive, song2025peeled}, hierarchical knowledge \cite{he2023language, sun2024large, kang2024researcharena, jiang2025hibench}, ontology reasoning \cite{he2023language, wang2024can}, ontology matching \cite{shvaiko2011ontology, kolyvakis2018deepalignment, kolyvakis2018biomedical, iyer-etal-2021-veealign, ibrahim2023toward}, and ontology learning \cite{jiang2010crctol, babaei2023llms4ol, e2eol2024lo, li2024ontology}. However, most existing datasets and benchmarks focus on only one or two specific aspects of ontologies, and few are designed specifically for LLMs with appropriate question-answering or generation formats. This limitation underscores the need for a comprehensive benchmark that covers a wide range of ontologies, domains, and tasks.

\section{OntoURL}

\subsection{Design Principle\label{sec:taxonomy}}

\textsc{OntoURL} is designed as an evaluation benchmark to systematically assess the multi-dimensional capabilities of LLMs within domain-specific ontologies. It serves two primary purposes: supporting ontology practitioners in selecting appropriate LLMs for ontology-related applications, and providing LLM researchers with insights into models' conceptual, hierarchical, reasoning and generative capabilities in ontological contexts.

\begin{figure}
    \centering
    \includegraphics[width=\linewidth]{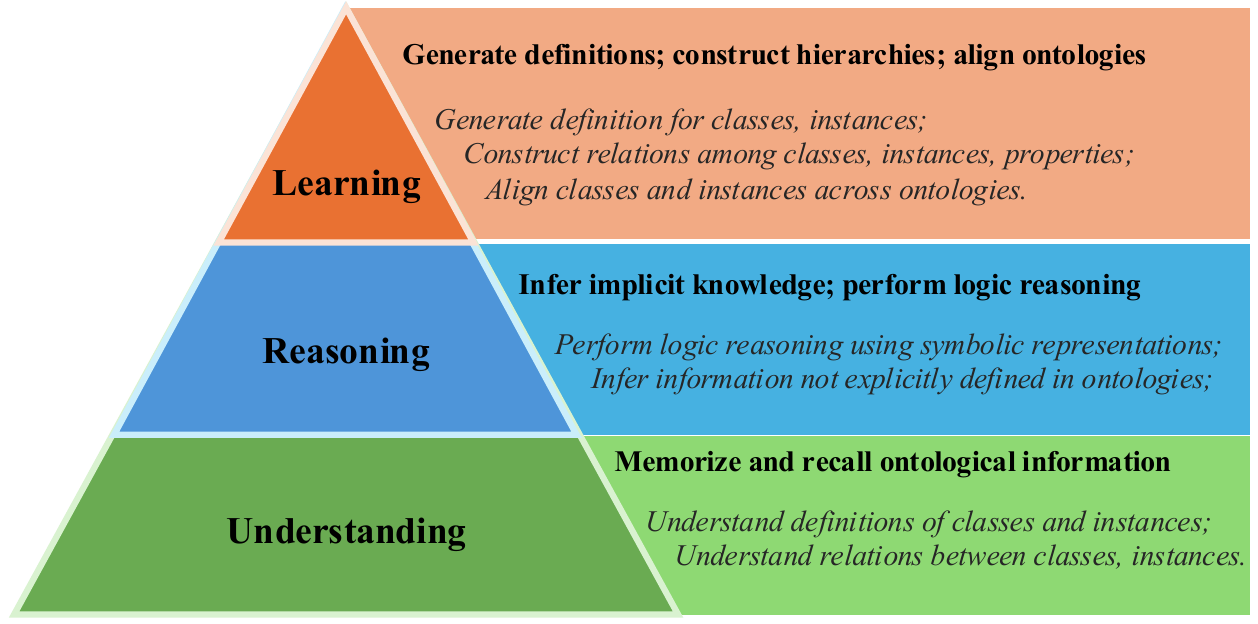}
    \caption{The taxonomy of LLM ontological capabilities, inspired by Bloom’s taxonomy. Each capability is positioned within a triangular structure and briefly explained on the right.}
    \label{fig:taxonomy}
\end{figure}

While considerable research has explored interactions between LLMs and ontologies, few studies have provided a systematic classification of the underlying capabilities. Drawing inspiration from Bloom's Taxonomy of educational objectives \cite{bloom1956taxonomy}, we introduce a three-level taxonomy for ontological capabilities for LLMs---\textbf{understanding}, \textbf{reasoning} and \textbf{learning} (Figure~\ref{fig:taxonomy}). 

\paragraph{Ontological Understanding} This represents the most fundamental ontological level and is thus placed at the base of the triangle in Figure~\ref{fig:taxonomy}. It encompasses the memorization, recall, and comprehension of explicitly defined ontology knowledge. For example, retrieving the definition of the concept ``calyx'' in the Plant Ontology, identifying its superclass and subsumption relationships, and recognizing its associated properties and instances.

\vspace{-3pt}
\paragraph{Ontological Reasoning} This capability builds upon ontological understanding and is positioned above it in the triangle of Figure~\ref{fig:taxonomy}. It involves inferring implicit knowledge that is not explicitly defined within an ontology. Structured ontologies often encode rich hierarchical relationships from which additional facts can be logically deduced. For example, the Plant Ontology states that ``testa'' is a subclass of ``seed coat'' ($\texttt{testa} \sqsubseteq \texttt{seed\ coat}$), ``seed coat'' is a subclass of ``seed'' ($\texttt{seed\ coat} \sqsubseteq \texttt{seed}$), and ``seed'' is a subclass of ``plant embryo'' ($\texttt{seed} \sqsubseteq \texttt{plant\ embryo}$). From these axioms, it follows that ``testa'' is also a subclass of ``plant embryo'' ($\texttt{testa} \sqsubseteq \texttt{plant\ embryo}$). We classify ontological reasoning as the ability to infer such implicit knowledge through reasoning process such as logical deduction.
\vspace{-3pt}
\paragraph{Ontological Learning} This capability represents the highest level in our taxonomy and is placed at the top of the triangle in Figure~\ref{fig:taxonomy}. It primarily concerns the process of constructing ontologies. Traditional ontology learning tasks have largely focused on generating hierarchical structures, while often neglecting other essential structural components like properties and instances. Therefore, we propose that ontological learning should encompass multiple dimensions: the generation of class definitions, the construction of class hierarchies, and the integration of properties and their constraints. In addition, ontology alignment---ensuring consistency across multiple ontologies by identifying and mapping equivalent concepts---is a critical aspect and is thus also considered part of this capability.

\subsection{Data Collection and Processing\label{sec:data_creation}}

\paragraph{Data Sources} \textsc{OntoURL} draws on 40 expert-created, open-source ontologies spanning a broad range of 8 different domains, including (1) sciences; (2) health and medicine; (3) business and finance; (4) earth and environment; (5) arts, media and entertainment; (6) food and agriculture; (7) human and society; and (8) the legal domain. All ontologies are provided in RDF \cite{miller1998introduction} or OWL \cite{mcguinness2004owl} format. 
While open-domain ontologies such as DBpedia \cite{auer2007dbpedia} are available, we focus on domain-specific ontologies due to their greater depth, consistency, and formal structure. In cases where multiple ontologies exist within the same domain, we address their heterogeneity by designing prompts tailored to each ontology.

\vspace{-3pt}

\paragraph{Data Processing} 

\begin{figure*}
    \centering
    \includegraphics[width=\linewidth]{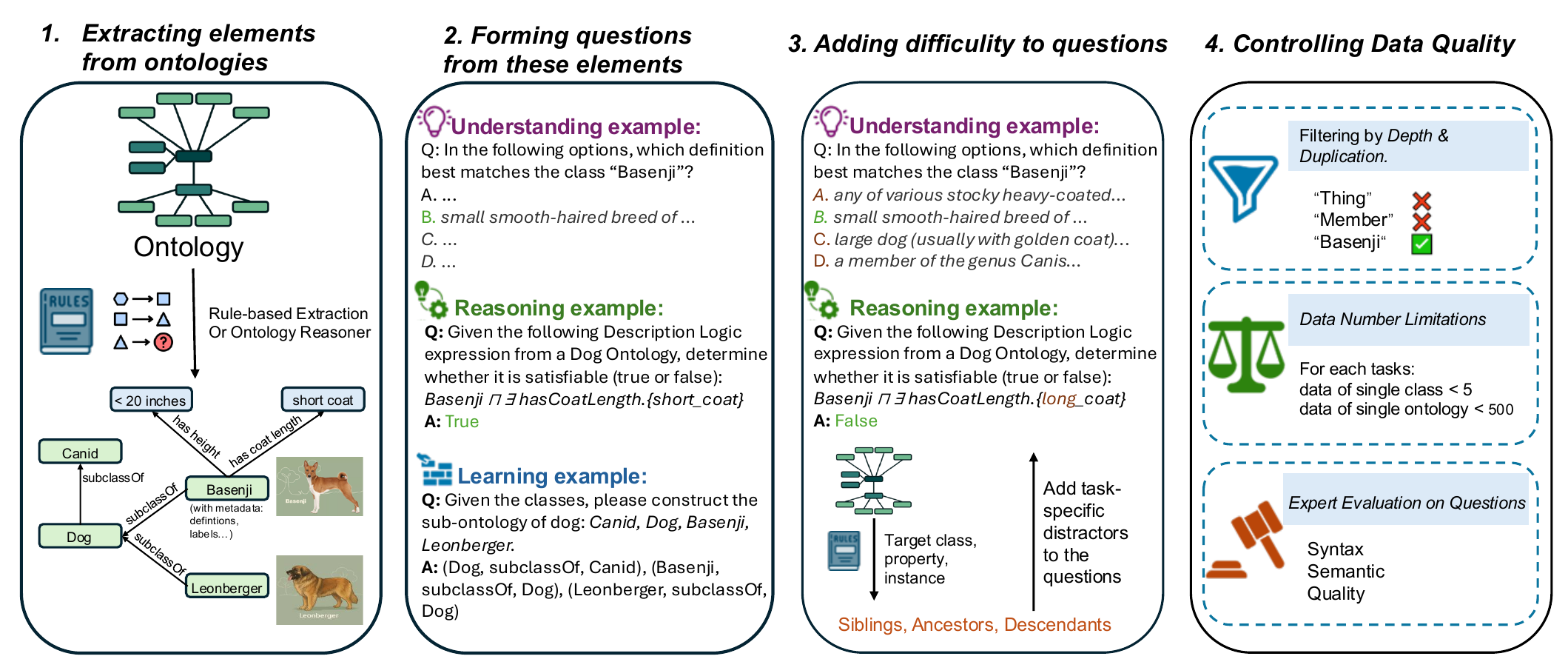}
    \caption{The pipeline of \textsc{OntoURL} construction: (1) elements are extracted from ontologies using rule-based extraction (understanding and learning tasks) and an ontology reasoner \textit{HermiT} (reasoning tasks); (2) the extracted elements are transformed into natural language questions; (3) distractors are added to form multiple-choice questions; and (4) the generated data is filtered and evaluated.}
    \label{fig:pipeline}
\end{figure*}

After collecting the ontologies, we apply a four-step pipeline to create multiple-choice, true/false question and open-ended questions, as illustrated in Figure~\ref{fig:pipeline}.

First, we extract task-relevant entities, including classes, instances, properties and their associated semantic details (e.g., definitions, relationships, and range) from each ontology. The sub-ontology of ``Basenji'' in the first step of Figure~\ref{fig:pipeline} presents the most involved elements in this extraction process. Particularly, an ontology reasoner is required to derive implicit relations for reasoning (e.g. the relation between ``Basenji'' and ``Canid''). 

Next, based on the extracted information, we construct natural language questions targeting different capabilities. Examples of questions for understanding, reasoning, and learning are shown in the second step of Figure~\ref{fig:pipeline}. 

For the multiple-choice questions, we generate answer options by selecting semantically plausible and structurally relevant distractors (e.g., ancestors, siblings, or children, as shown in the understanding example in the third step of Figure~\ref{fig:pipeline}). For true/false questions, we incorporate the distractors directly into the statements, as demonstrated in the reasoning example.

After question generation, we apply several filtering and balancing strategies to ensure quality and diversity: (a) To avoid over-representing abstract concepts, we assign sampling probabilities based on class depth (distance from the root) and sample data according to these probabilities. (b) Questions for the same classes are de-duplicated. For instance, as shown in Figure~\ref{fig:pipeline}, a class like ``Dog'' has multiple subclasses (e.g., ``Basenji'', ``Leonberger''), but only one question about the subsumptions of "Dog" will be retained. (c) We limit the number of questions per class (they may from different ontologies) to a maximum of five per task. (d) We set the total number of questions per ontology-task pair at 500 to prevent any single ontology from disproportionately influencing the evaluation results.

Finally, we perform an human verification to ensure the data quality in three key dimensions: (1) \textbf{Syntax}, to verify fluentness and grammatical correctness; (2) \textbf{Semantics}, to confirm the correctness of the answer; and (3) \textbf{Quality} (for multiple-choice questions), to assess whether the distractors are well-designed and appropriately challenging. Each expert reviewed 20\% of the data for each task, with a 5\% overlap between annotators to allow cross-verification of annotation consistency. This confirms the high quality of the benchmark, with over 95\% of questions rated as syntactically and semantically correct, and inter-annotator agreement exceeding 0.85 (Fleiss' $\kappa$, \citealp{landis1977measurement}) across evaluation dimensions (detailed guidelines and results are provided in Appendix~\ref{app:annotation}).

\subsection{Task Definition}

\begin{table*}
\centering 
\small
\renewcommand{\arraystretch}{0.85}%
\resizebox{\linewidth}{!}{
\begin{tabular}{lllllr}
\toprule
Capability    & ID        & Task Description & Question Type         & Metric & Sample Size \\ \midrule
\multirow{5}{*}{Understanding} & U1  & class definition understanding   & MCQ           & Accuracy &  9,151  \\
         & U2  & class relation understanding     & MCQ           & Accuracy &  9,201   \\
         & U3  & property domain understanding    & MCQ           & Accuracy &  375   \\
         & U4  & instance class understanding     & MCQ           & Accuracy &  2,475   \\
         & U5  & instance definition understanding  & MCQ         & Accuracy &  3,814  \\\midrule
\multirow{6}{*}{Reasoning}     & R1  & inferred relation reasoning         & MCQ        & Accuracy &  8,208   \\
         & R2  & constraint reasoning               & MCQ         & Accuracy &  6,956   \\
         & R3  & instance class reasoning           & MCQ         & Accuracy &  3,793   \\
         & R4  & swrl-based logic reasoning         & MCQ         & Accuracy &  6,517   \\
         & R5  & description logic reasoning        & T/FQ         & Accuracy &  882   \\ \midrule
\multirow{5}{*}{Learning}      & L1  & class definition generation         & Generation        & BERTScore & 2,936    \\
         & L2  & class hierarchy construction       & Generation         & Triple-F1 &  952   \\
         & L3  & property relation construction      & Generation        & Triple-F1 &  256    \\
         & L4  & constraint construction            & Generation         & Triple-F1 &  643     \\
         & L5  & ontology alignment                 & Generation         & Tuple-F1  &  1,149     \\ \bottomrule
\end{tabular}}
\caption{Overview of 15 tasks for evaluating ontological understanding, reasoning, and learning capabilities. Note: MCQ = Multiple-Choice Question; T/FQ = True/False Question.\label{tab:task_details}}
\end{table*}

We developed a series of tasks corresponding to the three ontological capabilities. An overview of these tasks is provided in Table~\ref{tab:task_details}.

For \textbf{Understanding} capability, tasks evaluate a LLM’s ability to comprehend explicitly defined ontological elements, including class definitions (U1), class relationships (U2), property domains (U3), instance classifications (U4), and instance definitions (U5).

The \textbf{Reasoning} capability increases complexity by requiring inference over implicit knowledge not explicitly presented in the ontology. Inferred relation reasoning (R1) extends task U2 by shifting from explicit to inferred class relationships. Similarly, constraint reasoning (R2) and instance class reasoning (R3) are inference-based counterparts of tasks U3 and U4, respectively. Tasks R4 and R5 introduce more advanced logical inference: SWRL-based logic reasoning (R4) involves reasoning over rules defined in the Semantic Web Rule Language \cite[SWRL]{horrocks2004swrl}, encompassing conjunctions, property chains, and conditional implications. Description logic reasoning (R5) focuses on reasoning with description logic, where models must interpret formal expressions and perform deductive inference over constructs such as $\exists$, $\forall$, $\sqcap$, and numerical restrictions (e.g., $\geq n$, $\leq m$).

The \textbf{Learning} capability tasks are generative and typically involve longer and more complex input contexts, making them more challenging than multiple-choice tasks. Class definition generation (L1) corresponds to task U1, requiring models to generate class definitions based on names and related information. Class hierarchy construction (L2) and property relation construction (L3) align with task U2. Constraint construction (L4) builds on task U3 by requiring models to generate constraints. Ontology alignment (L5) evaluates whether models can align semantically equivalent classes and instances across two ontologies.

\subsection{Benchmark Statistics}

\begin{figure}[!t]
    \centering
    \includegraphics[width=\linewidth]{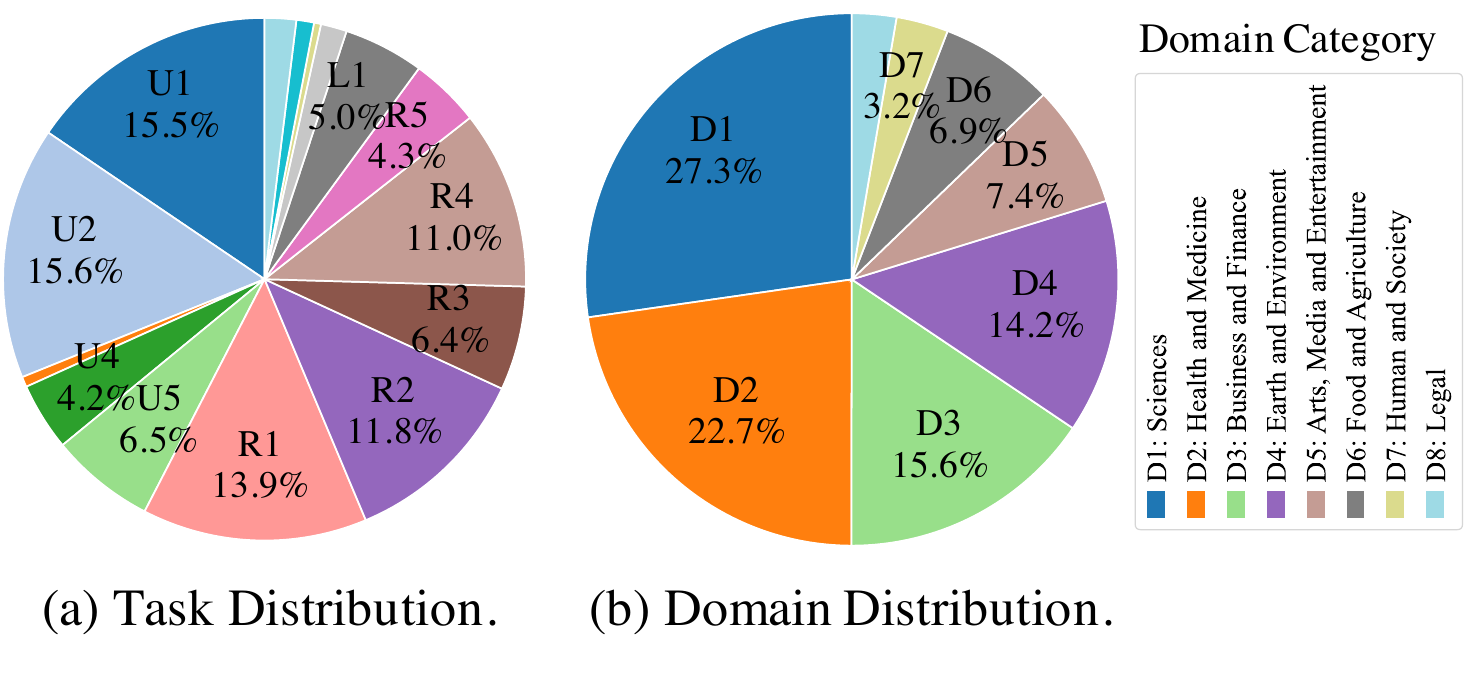}
    \caption{Question distribution of \textsc{OntoURL} tasks and domains. Additional statistics, such as average lengths of questions, options, and answers, are provided in Appendix~\ref{app:statistics}.}
    \label{fig:overall_distribution}
\end{figure}

Figure~\ref{fig:overall_distribution} presents the distribution of questions across tasks and domains in the \textsc{OntoURL} benchmark. As shown in Figure~\ref{fig:overall_distribution}a, Tasks U2 (Class Relation Understanding, 15.6\%), U1 (Class Definition Understanding, 15.5\%), and R1 (Inferred Relation Reasoning, 13.9\%) are the most prevalent in our benchmark. Conversely, ontology learning (L1-L5) and property-related tasks (U3, R2, L3, L4) constitute a smaller portion of the dataset. This distribution stems from two primary factors. First, most ontological classes contain superclasses and definitions, enabling the generation of more questions for tasks U1, U2, and R1. In contrast, properties and their associated constraints are not consistently provided across all ontologies, resulting in fewer questions for property-related tasks. Additionally, the ontology learning tasks were significantly reduced during the filtering process, which systematically eliminated overlapping sub-ontologies to ensure data quality and prevent redundancy.


The domain distribution (Figure~\ref{fig:overall_distribution}b) is directly related to the quantity and scale of the collected ontologies. The Sciences domain represents the largest portion (28.4\%), as it encompasses 8 ontologies, including extremely large resources like Gene Ontology \cite{ashburner2000gene, gene2023gene} and Cell Ontology \cite{diehl2016cell}. The Health \& Medicine domain follows as the second largest (22.7\%), while the Legal domain represents the smallest share (2.1\%), comprising only four relatively small ontologies.

\vspace{-5pt}
\section{Evaluation}


We evaluate LLMs under both zero-shot and few-shot settings across all 15 tasks in \textsc{OntoURL}. For the zero-shot scenario, the input to the LLMs consists solely of task instructions, questions, and answer options (where applicable). In the few-shot setting, we provide two or four carefully selected examples for each task to demonstrate the expected reasoning pattern and output format. As shown in Table~\ref{tab:task_details}, we use task-appropriate metrics: Accuracy for multiple-choice and true/false questions (tasks U1-U5, R1-R5), BERTScore F1 \cite{bertscore} for text generation (task L1), and F1 score for structured outputs such as triples or tuples (tasks L2-L5). We apply regular expressions to extract valid triples or tuples from the model's responses to mitigate the impact of irrelevant text. The hyperparameters and configurations are detailed in Appendix~\ref{app:experiment}.

\subsection{Evaluated Models}
We evaluate a diverse set of 20 language models, which can be categorized into three groups. \textbf{General-purpose LLMs} include 14 widely used open-source models across various parameter scales: Qwen2.5-(3B, 7B, 14B, 32B, 72B)~\cite{qwen2025qwen25technicalreport}, QWQ-32B, Phi4-4B~\cite{abdin2024phi4technicalreport}, LLaMA3.1-8B, LLaMA3.3-70B~\cite{grattafiori2024llama3herdmodels}, Aya-Expanse-(8B, 32B), InternLM3-8B, Mistral-8B, and Mistral-small. \textbf{Ontology-trained LLMs} comprise two task-specialized models—Ollm-wiki and Ollm-arxiv~\cite{e2eol2024lo}—which are fine-tuned from Mistral-7B~\cite{jiang2023mistral7b} on Wikipedia category and arXiv taxonomy data, respectively. \textbf{Domain-specialized LLMs} include SaulLM-7B (legal domain)~\cite{colombo2024saullm7bpioneeringlargelanguage}, BioMistral-7B (sciences)~\cite{labrak2024biomistralcollectionopensourcepretrained}, OpenBioLLM-8B (biomedicine), and Finance-Chat-7B (finance)~\cite{cheng2024adaptinglargelanguagemodels}. These models are included to provide a complementary perspective by assessing how domain-specific pretraining affects performance on ontological tasks. Links to all model repositories are provided in Appendix~\ref{app:model_url}.

\subsection{Experimental Results and Analysis}

We present the performance of models in Table~\ref{tab:zero-shot-performance} under the zero-shot setting.
In the following analysis, we discuss the results from three perspectives: model performance, ontological capabilities, and performance on specific domains. 

\begin{table*}
\small
\centering
\setlength{\tabcolsep}{3.5pt}
\renewcommand{\arraystretch}{0.88}%
\begin{tabular}{l | *{5}{r} | *{5}{r} | *{5}{r}}
\toprule
\textbf{Model}
& \multicolumn{5}{c|}{\textbf{Understanding} \scriptsize{(Acc.)}} 
& \multicolumn{5}{c|}{\textbf{Reasoning} \scriptsize{(Acc.)}} 
& \multicolumn{5}{c}{\textbf{Learning} \scriptsize{(BERTScore, F1)}} \\
\cmidrule(lr){2-6} \cmidrule(lr){7-11} \cmidrule(lr){12-16}
 & U1 & U2 & U3 & U4 & U5 & R1 & R2 & R3 & R4 & R5 & L1 & L2 & L3 & L4 & L5 \\
\midrule
\cellcolor[gray]{0.9} 3-4B &   &   &   &   &   &   &   &   &   &   &   &   &   &      &       \\
Qwen2.5-3B        & 77.8       & 86.3       & 80.3       & 85.8       & 77.8       & 81.5       & 74.9       & 65.7       & 62.5       & 67.9       & 79.8        & 0.1        & 0.0        & 0.2 & 6.7  \\
Phi4-4B &  77.5      & 91.1       & 75.5       & 87.2       & 78.8       & 80.2       & 80.7       & 63.5       & 59.1       & 51.1       & 82.4        & 0.1        & 0.0        & 0.0     & 0.1  \\ \midrule
\cellcolor[gray]{0.9}7-8B  &   &   &   &   &   &   &   &   &   &   &   &   &   &      &       \\
Qwen2.5-7B \good        & 83.1       & 90.6       & 77.6       & 90.1       & 83.6       & 87.6       & 88.2       & 73.9       & 66.0       & 68.6       & 79.8        & 0.4        & 0.1        & 0.3 & 16.2 \\
Ollm-wiki-7B      & 74.3       & 84.5       & 67.2       & 81.4       & 77.0       & 65.2       & 83.3       & 53.4       & 57.3       & 59.3        & 79.0        & 0.1        & 0.0        & 0.2 & 0.1  \\
Ollm-arxiv-7B     & 74.1       & 84.4       & 67.5       & 81.5       & 77.0       & 64.2       & 82.8       & 53.1       & 56.6       & 58.4        & 79.1        & 0.1        & 0.0        &  0.0    & 8.3  \\
LLaMA3.1-8B       & 79.8       & 87.4       & 74.9       & 88.4       & 81.1       & 79.8       & 84.2       & 72.3       & 62.2       & 68.9       & 79.4        & 0.1        & 0.0        &  0.1    & 15.3 \\
Ministral-8B      & 78.9 & 88.8 & 62.4 & 83.9 & 79.5 & 81.0 & 88.4 & 60.1 & 62.4 & 52.7 & \textbf{82.6} & 0.1 & 0.0 & 0.1 & 16.4       \\
Internlm3-8B      & 83.1       & 90.9       & 72.0       & 88.9       & 82.4       & 88.5       & 90.5       & 73.8       & 67.2       & 62.9       & 79.7        & 0.2        & 0.0        &   0.4   & 12.0 \\
Aya-8B  & 77.1       & 85.8       & 62.4       & 83.8       & 77.9       & 73.0       & 78.0       & 62.6       & 57.4       & 63.6       & 80.5        & 0.1        & 0.0         & 0.0 & 6.3  \\ \midrule
\cellcolor[gray]{0.9}14-32B&   &   &   &   &   &   &   &   &   &   &   &   &   &      &       \\
Qwen2.5-14B       & 86.6       & 92.0       & 75.5       & 91.4       & 85.8       & 89.6       & 94.0       & 76.4       & 71.2       & 63.6       & 79.9       & 0.1        & 0.1        & 1.0 & 19.5 \\
Mistral-22B       & 83.9       & 90.4       & 69.6       & 88.6       & 84.4       & 86.3       & 86.9       & 69.3       & 64.0       & 54.3       & 80.1        & 0.1        & 0.0        & 0.8 & 15.8 \\
Qwen2.5-32B \good       & 88.0       & 90.6       & 81.9       & 91.2       & 87.2       & 89.7       & \textbf{95.5}       & 76.8       & 72.4       & 68.4       & 80.0        & \textbf{1.6}        & 0.1        & 1.5 & 20.3 \\
QwQ-32B & 82.2       & 89.6       & 77.1       & 88.9       & 81.5       & 84.0       & 92.5       & 70.8       & 60.6       & 63.4       & 79.4        & 1.1        & \textbf{0.2}        & 1.0 & 18.0 \\
Aya-32B & 81.2       & 90.5       & 61.6       & 89.7       & 82.3       & 85.5       & 83.1       & 70.3       & 66.0       & \textbf{68.8}       & 79.3        & 0.1        & 0.1        & 0.5 & 19.3 \\ \midrule
\cellcolor[gray]{0.9}70-72B&   &   &   &   &   &   &   &   &   &   &   &   &   &      &       \\
LLaMA3.3-70B      & 88.0       & \textbf{94.1}       & 76.8       & 91.8       & \textbf{90.0}       & 91.9       & 92.9      & 76.8       & 70.9       & 64.2       & 80.0        & 0.1        & 0.0        & 0.7 & 20.2 \\
Qwen2.5-72B \best       & \textbf{89.1}       & 92.6       & \textbf{84.3}       & \textbf{92.6}       & 89.4       & \textbf{92.1}       & 93.4       & \textbf{77.5}       & \textbf{75.6}       & 68.4       & 79.9        & 0.1        & 0.0        & 1.0 & \textbf{21.6} \\ \bottomrule
\end{tabular}
\caption{Main results (\%) of 16 LLMs (grouped by size) under the zero-shot setting. \best\ indicates the best-performing model overall, while \good\ denotes the best-performing model within its size category.}
\label{tab:zero-shot-performance}
\end{table*}

\subsubsection{Performance of LLMs} 

The two largest models, LLaMA3.3-70B and Qwen2.5-72B, consistently achieve the best performance. Notably, the Qwen architecture shows robust results at all sizes, outperforming other architectures of comparable scale. In contrast, Ollm, which is specifically trained for ontology construction, performs relatively poorly, likely due to its specialization in hierarchical generation rather than general understanding or reasoning.

Model scale correlates strongly with performance, especially in understanding and reasoning. For instance, Qwen2.5-72B achieves top scores of 92.6 on U4, 93.4 on R2, and 21.6 F1-score on L5. Similarly, LLaMA3.3-70B scores 91.8 on U4, 92.9 on R2, and 20.2 on L5. This pattern is even more pronounced within the same architecture: across Qwen, Mistral, Aya, and LLaMA, larger models consistently perform better.

\paragraph{Ontological Understanding of LLMs} 

Tasks U1 to U5 demonstrate that LLMs generally perform well on ontology understanding, particularly in recognizing hierarchical structures. This is reflected in the high accuracy on U2 (class relations) and U4 (instance classification), with scores ranging from 80\% to 94\%. In contrast, performance is less consistent on definition and property tasks. U1 (class definition), U3 (property domain), and U5 (instance definition) show notable gaps in certain models, for instance, Aya-8B scores only 77.1\%, 62.4\%, and 77.9\% on these tasks.



\paragraph{Ontological Reasoning of LLMs} 

Reasoning tasks present greater challenges for LLMs than understanding tasks. Task R1 (inferred relation reasoning) is a difficult variant of task U2 (class relation understanding), requiring reasoning to identify class relationships not explicitly defined. As expected, models generally perform 3-4 percentage points worse on R1 than on U2, with the most dramatic decrease observed in Ollm-arxiv-7B (from 84.4\% to 64.2\%). Similarly, R3 (instance class reasoning) functions as the reasoning-intensive counterpart to U4. Performance on R3 (60-70\%) demonstrates a substantial decline compared to U4 (80-90\%). These results indicate that when reasoning across multiple relationships is required, performance deteriorates significantly.

Tasks involving logical expressions, such as R4 (SWRL-based reasoning) and R5 (description logic reasoning), are more difficult. Compared to natural language-based reasoning (R1, R2 and R3), model performance drops more dramatically when logical operators are involved, with scores ranging between 60\% and 75\%. Even the best-performing model, Qwen2.5-72B, achieves only 75.6\% and 68.8\% on these two tasks, respectively. This highlights a significant limitation of current LLMs: difficulty in executing precise symbolic reasoning over formally structured ontologies.

\paragraph{Ontological Learning of LLMs}
Although direct comparisons is limited due to different evaluation metrics, generation tasks are  shown to be more challenging, comparing with understanding and reasoning tasks. In L1 (class definition generation), models struggle to generate definitions, as evidenced by low BERTScore (typically below 10). This poor performance likely stems from the challenges of domain-specific definition generation, which requires not only describing the target class but also distinguishing it from adjacent concepts (e.g. its superclasses). Unlike human domain experts who possess comprehensive ontological perspectives, LLMs struggle to make such fine-grained semantic distinctions within specialized domains. 

Similar limitations appear in structural construction tasks. In L2 (class hierarchy construction), L3 (property relation construction), and L4 (constraint construction), models frequently fail to produce syntactically valid triples in zero-shot settings. Performance improves modestly in two- and four-shot settings, but output quality remains low, often featuring ill-defined or hallucinated relations. For L5 (ontology alignment), models achieve slightly higher F1 scores in the range of 10-20\%. But for practical ontology applications, the performance of the evaluated LLMs on all learning tasks remains substantially poor for reliable deployment.

\subsubsection{Analysis}

\paragraph{Domain-Specific Capabilities of LLMs}

\begin{figure}
  \centering
  \includegraphics[width=\linewidth]{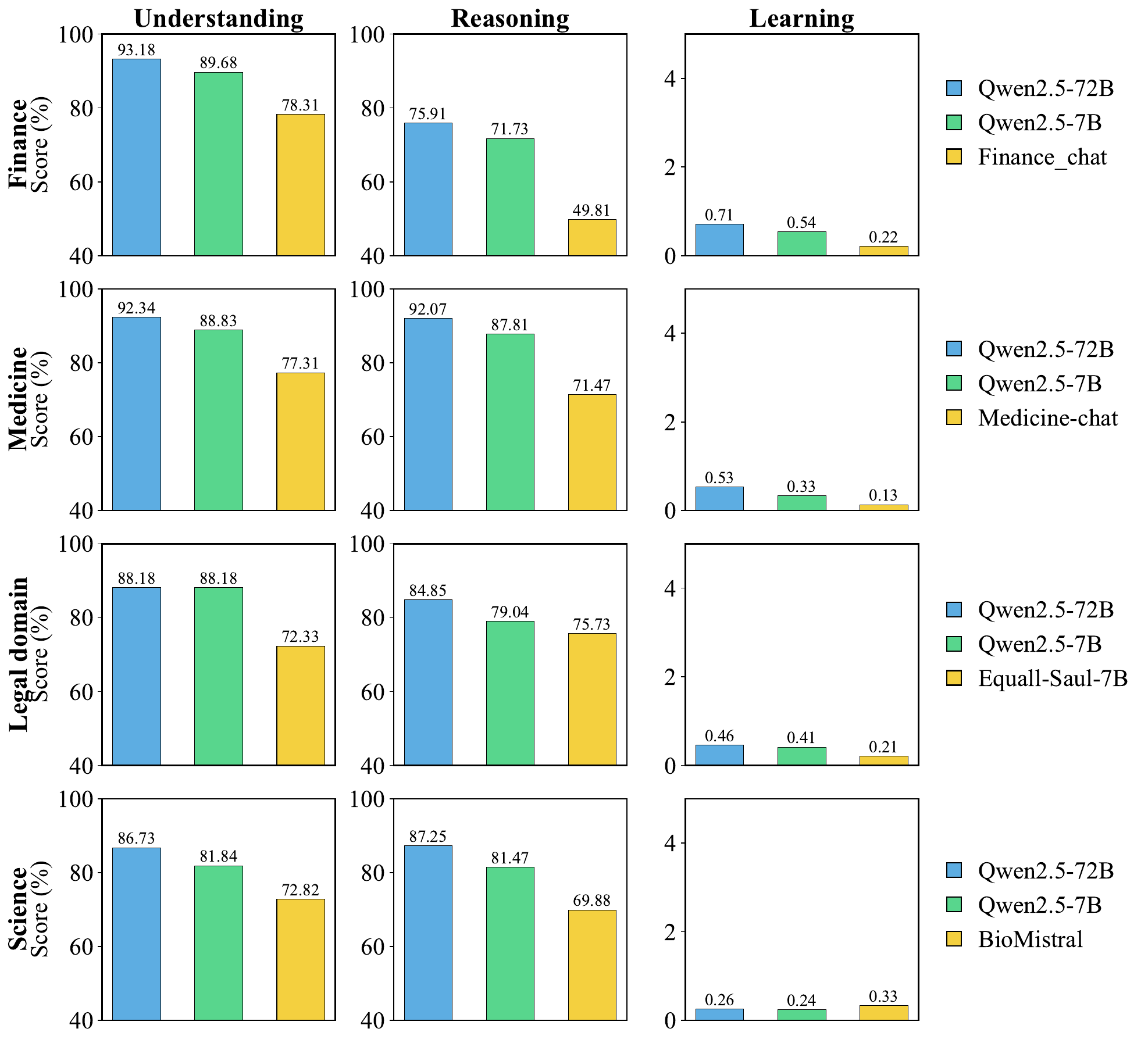}
  \caption{The performance of Qwen2.5-7B, 72B (green and blue) and four domain-specific LLMs (yellow) across four domains.}
  \label{fig:domain_performance}
\end{figure}

In addition to the task‐based evaluation, Figure \ref{fig:domain_performance} compares the performance of  two open-domain LLMs, and four domain‐specific LLMs (Finance-Chat-7B, OpenBioLLM-8B, BioMistral-7B, and SaulLM-7B). For the Sciences domain, we retained only the Biology‐related tasks. To simplify the computation of the mean scores for the learning task, we omitted the scores of L1.

As expected, the trends observed earlier hold as well: larger models consistently outperform smaller ones, and models struggle more on reasoning and learning than on understanding tasks. Comparing across domains, performance in Legal and the Sciences (Biology) lags behind the other two domains, with the gap most pronounced in Biology, which is likely a reflection of the greater complexity and specialized knowledge required.

In terms of model comparison, we find that the open-domain LLMs generally outperform their domain-specific counterparts, particularly on reasoning tasks. This supports our earlier observation on Ollm that fine-tuning for specific domains or tasks can erode a model’s generalization ability, leading to diminished performance when confronted with novel task formats.

\vspace{-3pt}

\begin{figure}
    \centering
    \includegraphics[width=\linewidth]{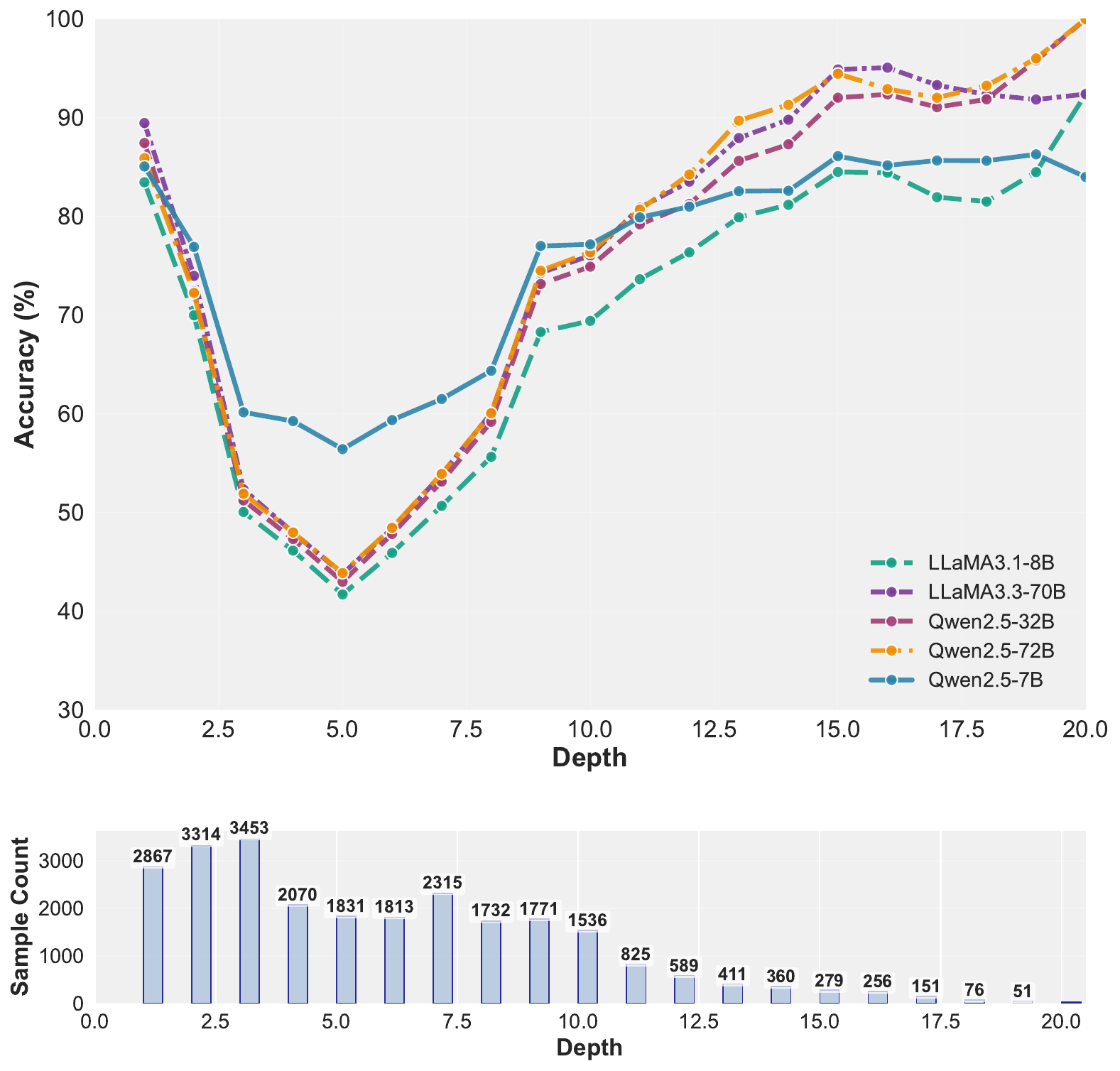}
    \caption{performance of five models with concept depth (0–20). The results aggregate tasks U1–U5 and R1–R3. Depths $>$ 20 are omitted due to scarce samples.}
    \label{fig:depth}
\end{figure}

\paragraph{Does Concept Depth matter?}
Concepts occupy different positions within the ontology and vary systematically in difficulty. As shown in Figure~\ref{fig:depth}, model accuracy exhibits a consistent U-shaped pattern across all five models: performance is high at shallow depths (1–3), drops sharply at intermediate depths (4–8), and recovers beyond depth 10. This trend persists despite reduced sample sizes at greater depths, indicating that the recovery is not an artifact of data volume. We hypothesize that intermediate-depth concepts present a fundamental challenge: they are too specific to benefit from broad generalization yet too abstract to be directly memorized from pretraining data. The recovery at deeper levels suggests that highly specific, leaf-node concepts may contain sufficient distinctive features for easier classification. These results reveal a structural limitation in how LLMs process concepts at different levels of abstraction.

\vspace{-1pt}

\paragraph{Does Few-shot Prompting help?}

We evaluated 0-shot, 2-shot, and 4-shot prompting across five models. Few-shot examples yield modest improvements on Understanding and Reasoning tasks: Qwen-72B improves from 89.6\% (0-shot) to 91.3\% (4-shot). The most substantial gains occur on Learning tasks: Qwen-72B increases from 7.7\% to 21.0\%, and Qwen-32B from 5.7\% to 19.1\%. These improvements scale with model size, suggesting that few-shot effectiveness depends on both task complexity and model capacity. Notably, LLaMA-8B exhibits a slight performance decline on Reasoning with 4-shot, which we attribute to context length limitations in smaller models when processing longer prompts.

\begin{figure}
    \centering
    \includegraphics[width=\linewidth]{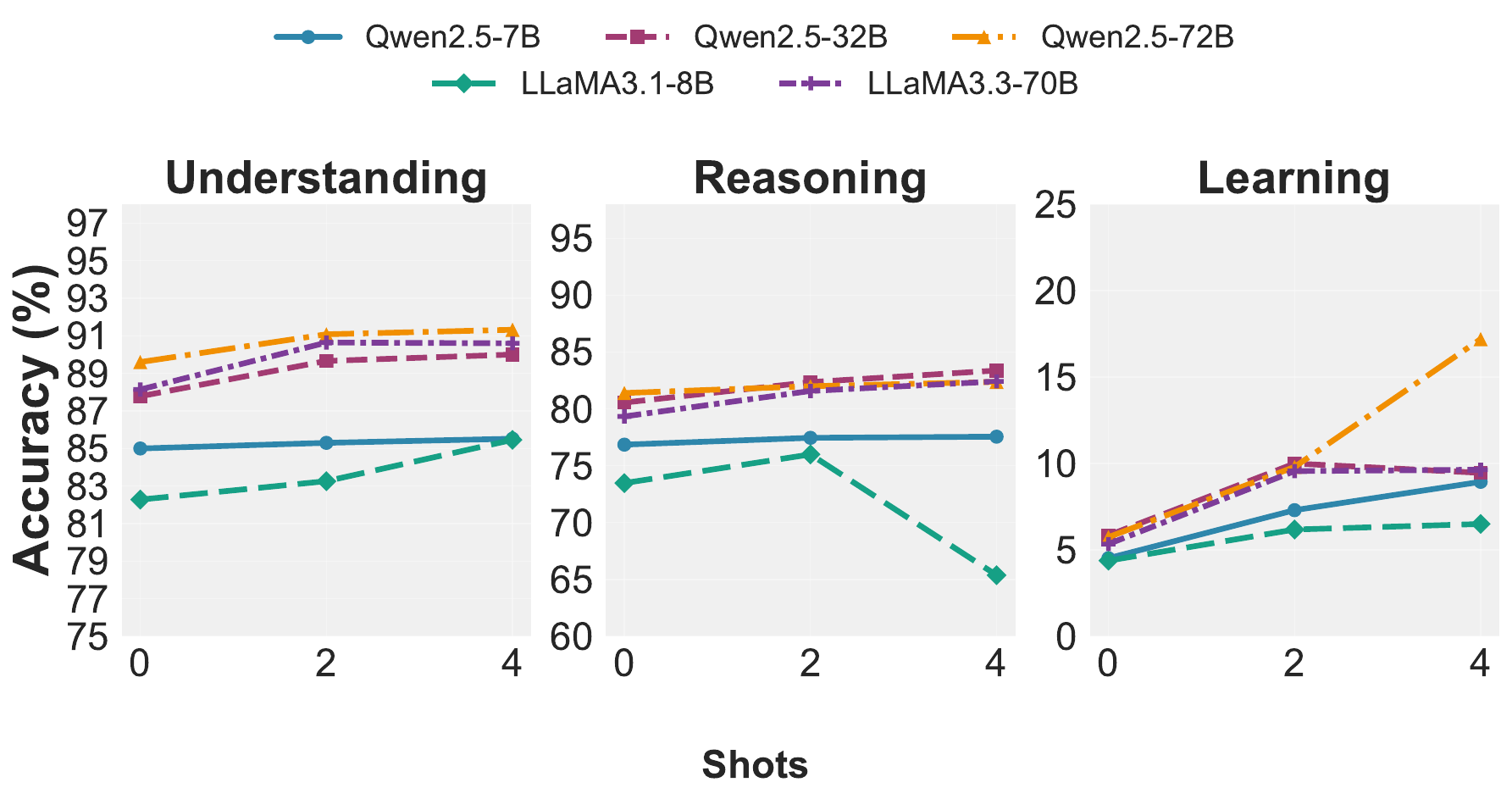}
    \caption{Performance of five models with 0, 2, and 4-shot prompting across three levels. Scores are aggregated within each task level. Full results are shown in Appendix~\ref{app:more_results}.}
    \label{fig:few_shot_res}
\end{figure}

\vspace{-1pt}

\begin{table*}
\centering
\footnotesize
\setlength{\tabcolsep}{5pt}%
\renewcommand{\arraystretch}{0.75}%
\begin{tabular}{@{}l *{15}{r}@{}}
\toprule
\multirow{2}{*}{\textbf{System}} & \multicolumn{5}{c}{\textbf{Understanding}} & \multicolumn{5}{c}{\textbf{Reasoning}} & \multicolumn{5}{c}{\textbf{Learning}} \\
\cmidrule(lr){2-6}\cmidrule(lr){7-11}\cmidrule(lr){12-16}
& U1 & U2 & U3 & U4 & U5 & R1 & R2 & R3 & R4 & R5 & L1 & L2 & L3 & L4 & L5 \\ \midrule Human1 & 55.0 & 60.0 & 65.0 & 65.0 & 65.0 & 75.0 & 70.0 & 75.0 & 65.0 & 70.0 & 3.0 & 20.1 & 51.0 & 49.0 & 51.7 \\ Human2 & 65.0 & 60.0 & 60.0 & 75.0 & 75.0 & 80.0 & 70.0 & 65.0 & 60.0 & 60.0 & 2.0 & 30.1 & 48.3 & 61.0 & 53.0 \\ Human3 & 60.0 & 60.0 & 62.5 & 70.0 & 70.0 & 77.5 & 70.0 & 70.0 & 62.5 & 65.0 & 2.5 & 25.2 & 49.5 & 55.0 & 52.5 \\ \midrule Qwen-7B & 83.1 & 90.6 & 77.6 & 90.1 & 83.6 & 87.6 & 88.2 & 73.9 & 66.0 & 68.6 & 5.6 & 0.4 & 0.1 & 0.3 & 16.2 \\ Qwen-32B & 88.0 & 90.6 & 81.9 & 91.2 & 87.2 & 89.7 & 95.5 & 76.8 & 72.4 & 68.4 & 5.7 & 1.6 & 0.1 & 1.5 & 20.3 \\ Qwen-72B & 89.1 & 92.6 & 84.3 & 92.6 & 89.4 & 92.1 & 93.4 & 77.5 & 75.6 & 68.4 & 6.0 & 0.1 & 0.0 & 1.0 & 21.6 \\ \bottomrule
\end{tabular}
\caption{Comparison of humans and LLMs. Human results are computed over 30 randomly selected questions per task, without the help of any tools; LLM results are reported on the full dataset.}
\label{tab:human}
\end{table*}

\paragraph{Does Chain of Thought (CoT) help?}

\begin{figure}
    \centering
    \includegraphics[width=\linewidth]{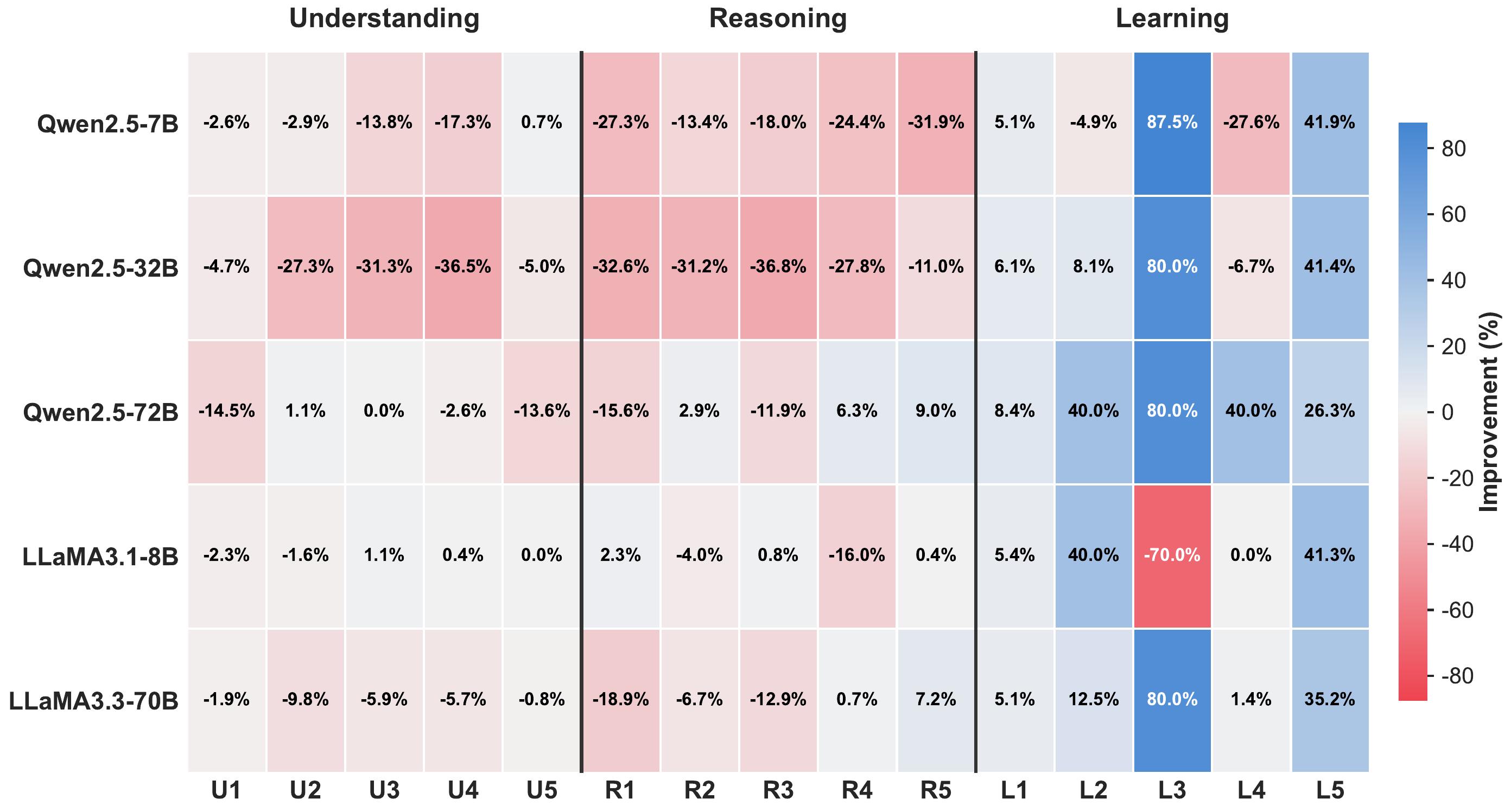}
    \caption{Performance impact of chain-of-thought prompting across five models, compared to standard prompting. Red indicates performance decreases, and blue indicates improvements. Full results are shown in Appendix~\ref{app:more_results}.}
    \label{fig:cot_res}
\end{figure}

We assess the impact of Chain-of-Thought prompting on  performance. As shown in Figure~\ref{fig:cot_res}, CoT prompting produces mixed results. On Understanding and Reasoning tasks, we observe modest to significant performance declines: Qwen2.5-7B drops from 83.1\% to 80.9\% on Understanding, while Qwen-72B shows larger decreases of 14.5pp and 15.6pp on Understanding and Reasoning, respectively. Conversely, Learning tasks demonstrate substantial improvements: Qwen-7B increases from 5.5\% to 12.4\%, Qwen-32B from 5.7\% to 20.2\%, and Qwen-72B from 6.0\% to 18.5\%.

These contrasting effects suggest that CoT's impact is task-dependent. We propose two explanations: (1) Understanding and Reasoning tasks in \textsc{OntoURL} require ontology-specific inference patterns that differ fundamentally from the mathematical and commonsense reasoning prevalent in pretraining corpora, limiting CoT transferability; (2) Learning tasks benefit from CoT's structured generation process, which helps models organize knowledge during content creation.

\paragraph{LLMs vs. Humans}

To contextualize LLM capabilities, we conducted a human evaluation with three participants holding PhD-level expertise. As shown in Table~\ref{tab:human}, LLMs substantially outperform humans on Understanding tasks (average gap: +22.9pp), where formal ontological concepts in specialized domains (e.g., medical, science) pose significant challenges. The LLM advantage narrows on Reasoning tasks (average gap: +10.6pp), suggesting these tasks rely more on general logical reasoning---a skill less dependent on wide knowledge. Notably, humans achieve superior performance on Learning tasks (average gap: -31.6pp), though both groups struggle with definition generation (Task L1: humans 2.5\%, LLMs 5.8\%). Overall, the results suggest that LLMs excel at leveraging broad ontological knowledge, but face challenges in tasks requiring creative synthesis and generation.

\section{Conclusion}

In this paper, we introduce \textsc{OntoURL}, a comprehensive benchmark for evaluating the ontological capabilities of LLMs. We propose a taxonomy encompassing three dimensions and develop a systematic pipeline for generating and validating questions. Evaluation results show that while LLMs exhibit strong performance in ontological understanding, they struggle with reasoning and learning. \textsc{OntoURL} further reveals several insights, including the particular difficulty of mid-level concepts, the impact of few-shot and chain-of-thought prompting, and performance differences between humans and LLMs. These findings highlight that despite rapid progress in LLM, significant challenges remain in handling symbolic ontological knowledge. We believe that \textsc{OntoURL} can be a valuable resource for both ontology practitioners and AI researchers, facilitating the evaluation, analysis and development of LLMs in ontology domain. Current limitations include restricted domain coverage, incomplete task types, and English-only scope---areas we will address as \textsc{OntoURL} evolves as a long-term project.

\newpage

\bibliography{tacl2021}
\bibliographystyle{acl_natbib}



\appendix

\section{Availability\label{app:availability}}
We provide access to the codebase for LLM experiments, evaluation tools, and all-related files (e.g. zero-, two- and four-shot prompts, chain-of-thought prompts, and model output files). \url{https://anonymous.4open.science/r/OntoURL_anonymous-44FD}

\section{Task Statistics\label{app:statistics}}

Table~\ref{tab:task_statistics} presents the statistics of each task and each domain in \textsc{OntoURL}. We list the number of samples for each task and domain and the average words of queries. 

\begin{table*}[ht]
\centering
\small
\renewcommand{\arraystretch}{0.9}%
\begin{tabular}{@{}lrrr@{}}
\toprule
\textbf{Task} & \textbf{Question} & \textbf{Option} & \textbf{Answer} \\
\midrule
U1 Class Definition Understanding            & 16.40 & 96.94 & - \\
U2 Class Relation Understanding              & 18.62 & 16.17 & - \\
U3 Property Domain Understanding             & 21.46 & 13.13 & - \\
U4 Instance Class Understanding              & 18.22 & 12.78 & - \\
U5 Instance Definition Understanding         & 19.55 & 85.27 & - \\\midrule
R1 Inferred Relation Reasoning               & 19.35 & 14.98 & - \\
R2 Constraint Reasoning                      & 45.72 & 12.36 & - \\
R3 Instance Class Reasoning                  & 19.39 & 12.57 & - \\
R4 SWRL-Based Logic Reasoning                & 18.61 & 18.06 & - \\
R5 Description Logic Reasoning               & 23.97 & -  & - \\\midrule
L1 Class Definition Generation               & 17.13 & -  & 25.87 \\
L2 Class Hierarchy Construction              & 241.91 & - & 93.41 \\
L3 Property Relation Construction            & 304.20 & - & 48.67 \\
L4 Constraint Construction                   & 323.95 & - & 76.76 \\
L5 Ontology Alignment                        & 534.46 & - & 135.19 \\
\bottomrule
\end{tabular}
\caption{Word counts across tasks in \textsc{OntoURL}, including questions, options, and answers. For multiple-choice questions, answer lengths are not considered; for true/false questions, option and answer lengths are excluded; and for generation tasks, option lengths are omitted.}
\label{tab:task_statistics}
\end{table*}

\begin{table*}[ht]
\centering
\small
\renewcommand{\arraystretch}{0.9}%
\begin{tabular}{@{}lrrr@{}}
\toprule
\textbf{Domain} & \textbf{Question} & \textbf{Option} & \textbf{Answer} \\
\midrule
Arts Media Entertainment  & 32.32  & 21.10 & 85.91 \\
Business Finance          & 31.98  & 39.82 & 45.01 \\
Earth Environment         & 25.54  & 26.81 & 39.17 \\
Food Agriculture          & 27.68  & 35.23 & 43.88 \\
Health Medicine           & 29.81  & 35.10 & 45.01 \\
Human Society             & 19.82  & 31.55 & 23.71 \\
Legal Domain              & 59.62  & 23.20 & 41.17 \\
Sciences                  & 30.38  & 41.73 & 46.94 \\
\bottomrule
\end{tabular}
\caption{Word counts across domains in \textsc{OntoURL}, including questions, options, and answers. For multiple-choice questions, answer lengths are not considered; for true/false questions, option and answer lengths are excluded; and for generation tasks, option lengths are omitted.}
\end{table*}

\section{Expert Verification\label{app:annotation}}

In Table~\ref{tab:quality_criteria}, we give the criteria which the three expert are asked to following during the verification of the data of \textsc{OntoURL}.

\begin{table*}[htbp]
\centering
\small
\renewcommand{\arraystretch}{0.9}
\resizebox{\textwidth}{!}{%
\begin{tabular}{c >{\raggedright\arraybackslash}p{4cm} >{\raggedright\arraybackslash}p{4cm} >{\raggedright\arraybackslash}p{4cm}}
\toprule
\textbf{Score} & \textbf{Syntax} & \textbf{Semantics} & \textbf{Distractor Quality} \\
\midrule
5 & No errors in spelling, grammar, punctuation, or casing; highly fluent. 
  & Question and answer align precisely; facts accurate and clear. 
  & All three distractors closely related, same category, similar difficulty. \\
4 & One or two minor errors (e.g., extra space, comma). 
  & Minor wording variations; correct answer clear. 
  & Most distractors relevant; one slightly off but functional. \\
3 & Several minor errors or few clear grammatical issues. 
  & Some ambiguity; answer deducible from context. 
  & Two or more distractors poorly related or uneven in difficulty. \\
2 & Multiple grammar errors hindering readability. 
  & Question and answer misaligned or need extra context. 
  & Most distractors irrelevant, too easy/hard, or confusing. \\
1 & Incomprehensible or meaningless text. 
  & Question and answer unrelated or incorrect. 
  & Distractors off-topic, erroneous, or incorrect in number. \\
\bottomrule
\end{tabular}}
\caption{Scoring criteria (1--5 scale) for evaluating syntax, semantics, and distractor quality in multiple-choice questions.}
\label{tab:quality_criteria}
\end{table*}

Table~\ref{tab:annotation-quality} reports the annotation results. Overall, the automatically generated data exhibit high syntactic and semantic quality, with strong agreement among annotators. This confirms the reliability of our rule-based pipeline and the clarity of the annotation protocol.

\begin{table*}[htbp]
\centering
\small
\renewcommand{\arraystretch}{0.9}%
\begin{tabular}{@{}lccc@{}}
\toprule
\textbf{Dimension} & \textbf{Acceptable Rate} & \textbf{IAA (Fleiss' $\kappa$)} & \textbf{Remarks} \\
\midrule
Syntax   & 97.8\% & 0.89 & Majority rated questions as fluent and grammatically correct \\
Semantics & 95.4\% & 0.85 & Most answers correctly reflect ontology knowledge \\
Quality & 92.6\% & 0.82 & Distractors generally appropriate and non-trivial \\
\bottomrule
\end{tabular}
\caption{Expert annotation results across dimensions and inter-annotator agreement (IAA). Each metric is reported as the percentage of instances rated as acceptable by at least two annotators.}
\label{tab:annotation-quality}
\end{table*}

\section{Details of Experiments\label{app:experiment}}

\paragraph{Hyperparameters}

All experiments were conducted using a unified inference framework. Table~\ref{tab:inference-config} summarizes the hardware, software, and inference hyperparameters used across all model evaluations.

\begin{table*}[htbp]
\centering
\small
\begin{tabular}{ll}
\toprule
\textbf{Category} & \textbf{Configuration} \\
\midrule
GPU & 4× NVIDIA H100 80GB \\
Batching & max\_batched\_tokens=8192 \\
Max Generation Length & 128 tokens (understanding and reasoning task), 512 tokens (learning task) \\
Temperature & 0.0 \\
Top-$p$ & 1.0 \\
Prompt Variants & Zero-shot, Two-shot, Four-shot \\
\bottomrule
\end{tabular}
\caption{Experimental setup for LLM inference.}
\label{tab:inference-config}
\end{table*}

\paragraph{Models\label{app:model_url}}

The models and their repositories are available in Table~\ref{tab:model_url}.

\begin{table*}[htbp]
\centering
\small
\renewcommand{\arraystretch}{0.9}%
\begin{tabular}{ll}
\toprule
Model         & Url \\ \midrule
Qwen2.5-3B    & \url{https://huggingface.co/Qwen/Qwen2.5-3B-Instruct}    \\
Phi4-4B       &  \url{https://huggingface.co/microsoft/Phi-4-mini-instruct}   \\
Qwen2.5-7B    &  \url{https://huggingface.co/Qwen/Qwen2.5-7B-Instruct}   \\
Ollm-wiki-7B  &  \url{https://huggingface.co/andylolu24/ollm-wikipedia}   \\
Ollm-arxiv-7B &  \url{https://huggingface.co/andylolu24/ollm-arxiv}  \\
Ministral-8B  &  \url{https://huggingface.co/mistralai/Ministral-8B-Instruct-2410}   \\
LLaMA3.1-8B   &  \url{https://huggingface.co/meta-llama/Llama-3.1-8B-Instruct}   \\
Internlm3-8B  &  \url{https://huggingface.co/internlm/internlm3-8b-instruct}  \\
Aya-8B        &   \url{https://huggingface.co/CohereLabs/aya-expanse-8b}  \\
Qwen2.5-14B   &   \url{https://huggingface.co/Qwen/Qwen2.5-14B-Instruct}  \\
Mistral-22B   &  \url{https://huggingface.co/mistralai/Mistral-Small-Instruct-2409}   \\
Qwen2.5-32B   &  \url{https://huggingface.co/Qwen/Qwen2.5-32B-Instruct}   \\
QwQ-32B       & \url{https://huggingface.co/Qwen/QwQ-32B}    \\
Aya-32B       &  \url{https://huggingface.co/CohereLabs/aya-expanse-32b}   \\
LLaMA3.3-70B  &  \url{https://huggingface.co/meta-llama/Llama-3.3-70B-Instruct}   \\
Qwen2.5-72B   &  \url{https://huggingface.co/Qwen/Qwen2.5-72B-Instruct}   \\
Finance-chat-7B & \url{https://huggingface.co/AdaptLLM/finance-chat} \\
Medicine-chat-7B & \url{https://huggingface.co/AdaptLLM/medicine-chat} \\
Equall-Saul-7B  & \url{https://huggingface.co/Equall/Saul-7B-Instruct-v1} \\
BioMistral & \url{https://huggingface.co/BioMistral/BioMistral-7B}\\\bottomrule
\end{tabular}
\caption{Models and its repositories.\label{tab:model_url}}
\end{table*}

\section{Additional Evaluation Result~\label{app:more_results}}

All the results for few-shot experiments and chain-of-thought experiments are shown in Table~\ref{tab:full_few_shot_results} and Table~\ref{tab:zero-shot-cot-performance}, respectively.

\begin{table*}
\centering
\tiny
\renewcommand{\arraystretch}{0.9}%
\resizebox{\textwidth}{!}{
\begin{tabular}{ll *{15}{r}}
\toprule
\multirow{2}{*}{Model} & \multirow{2}{*}{Shot} & \multicolumn{5}{c}{Understanding} & \multicolumn{5}{c}{Reasoning} & \multicolumn{5}{c}{Learning} \\
\cmidrule(lr){3-7} \cmidrule(lr){8-12} \cmidrule(lr){13-17}
& & U1 & U2 & U3 & U4 & U5 & R1 & R2 & R3 & R4 & R5 & L1 & L2 & L3 & L4 & L5 \\
\midrule
\rowcolor[gray]{0.9} \multicolumn{17}{l}{\textbf{3-4B Models}} \\
\multirow{3}{*}{Qwen2.5-3B} 
  & Zero & 77.8 & 86.3 & 80.3 & 85.8 & 77.8 & 81.5 & 74.9 & 65.7 & 62.5 & 67.9 & 79.8 & 0.1 & 0.0 & 0.2 & 6.7 \\
  & Two  & 81.6 & 88.3 & 76.3 & 86.9 & 79.2 & 79.4 & 91.4 & 63.7 & 68.0 & 68.7 & 80.5 & 8.5 & 0.0 & 1.0 & 6.5 \\
  & Four & 81.3 & 87.7 & 81.3 & 89.5 & 80.7 & 79.7 & 91.4 & 63.6 & 70.0 & 68.7 & 81.0 & 8.2 & 0.0 & 1.8 & 4.4 \\
\midrule
\multirow{3}{*}{Phi4-4B} 
  & Zero & 77.5 & 91.1 & 75.5 & 87.2 & 78.8 & 80.2 & 80.7 & 63.5 & 59.1 & 51.1 & 82.4 & 0.1 & 0.0 & 0.0 & 0.1 \\
  & Two  & 79.5 & 92.4 & 76.3 & 87.4 & 77.0 & 83.0 & 91.8 & 61.3 & 68.4 & 64.3 & 83.1 & 0.3 & 0.1 & 0.2 & 2.1 \\
  & Four & 78.9 & 93.2 & 81.3 & 87.6 & 78.9 & 82.5 & 91.3 & 64.7 & 70.9 & 63.6 & 83.6 & 0.3 & 0.2 & 0.9 & 1.7 \\
\midrule
\rowcolor[gray]{0.9} \multicolumn{17}{l}{\textbf{7-8B Models}} \\
\multirow{3}{*}{Qwen2.5-7B} 
  & Zero & 83.1 & 90.6 & 77.6 & 90.1 & 83.6 & 87.6 & 88.2 & 73.9 & 66.0 & 68.6 & 79.8 & 0.4 & 0.1 & 0.3 & 16.2 \\
  & Two  & 83.7 & 93.3 & 77.1 & 89.9 & 82.5 & 83.7 & 94.9 & 67.5 & 71.7 & 69.5 & 80.6 & 8.8 & 0.3 & 1.7 & 15.3 \\
  & Four & 83.8 & 93.1 & 76.3 & 90.8 & 83.6 & 82.0 & 95.3 & 67.7 & 73.8 & 69.0 & 81.1 & 15.0 & 0.6 & 2.0 & 10.2 \\
\midrule
\multirow{3}{*}{Ollm-wiki-7B} 
  & Zero & 74.3 & 84.5 & 67.2 & 81.4 & 77.0 & 65.2 & 83.3 & 53.4 & 57.3 & 9.3 & 79.0 & 0.1 & 0.0 & 0.2 & 0.1 \\
  & Two  & 77.5 & 87.6 & 71.2 & 85.9 & 75.0 & 64.9 & 86.9 & 53.1 & 68.0 & 64.0 & 79.8 & 13.3 & 0.2 & 1.2 & 3.3 \\
  & Four & 76.5 & 86.1 & 74.1 & 85.3 & 77.7 & 67.0 & 89.9 & 55.8 & 68.4 & 56.1 & 80.3 & 16.5 & 0.5 & 1.5 & 3.4 \\
\midrule
\multirow{3}{*}{Ollm-arxiv-7B} 
  & Zero & 74.1 & 84.4 & 67.5 & 81.5 & 77.0 & 64.2 & 82.8 & 53.1 & 56.6 & 8.4 & 79.1 & 0.1 & 0.0 & 0.0 & 8.3 \\
  & Two  & 77.4 & 87.5 & 70.4 & 85.9 & 74.9 & 64.0 & 87.0 & 52.9 & 67.9 & 62.4 & 79.9 & 13.5 & 0.2 & 1.6 & 3.2 \\
  & Four & 76.7 & 85.8 & 73.6 & 84.9 & 77.4 & 66.3 & 89.8 & 55.7 & 68.4 & 55.9 & 80.4 & 16.1 & 0.5 & 1.8 & 3.8 \\
\midrule
\multirow{3}{*}{Ministral-8B} 
  & Zero & 78.9 & 88.8 & 62.4 & 83.9 & 79.5 & 81.0 & 88.5 & 60.1 & 62.4 & 52.7 & 82.6 & 0.1 & 0.0 & 0.1 & 16.4 \\
  & Two  & 81.5 & 93.6 & 68.3 & 89.8 & 80.4 & 81.7 & 89.0 & 61.2 & 67.7 & 68.6 & 83.4 & 23.2 & 0.1 & 1.9 & 10.2 \\
  & Four & 81.9 & 93.7 & 96.5 & 89.1 & 81.8 & 80.0 & 90.0 & 63.3 & 71.4 & 70.1 & 83.9 & 22.6 & 0.3 & 2.2 & 9.8 \\
\midrule
\multirow{3}{*}{Llama3.1-8B} 
  & Zero & 79.8 & 87.2 & 74.9 & 88.4 & 81.1 & 79.8 & 84.2 & 72.3 & 62.2 & 68.9 & 79.4 & 0.1 & 0.0 & 0.1 & 15.3 \\
  & Two  & 82.6 & 91.5 & 70.4 & 89.9 & 81.9 & 85.2 & 92.5 & 64.6 & 69.4 & 68.3 & 80.2 & 13.3 & 0.1 & 2.5 & 10.5 \\
  & Four & 83.2 & 91.5 & 78.7 & 90.8 & 83.1 & 84.5 & 93.4 & 5.9 & 75.1 & 68.0 & 80.7 & 15.9 & 0.4 & 2.1 & 9.7 \\
\midrule
\multirow{3}{*}{Internlm3-8B} 
  & Zero & 83.1 & 90.9 & 72.0 & 88.9 & 82.4 & 88.5 & 90.5 & 73.8 & 67.2 & 62.9 & 79.7 & 0.2 & 0.0 & 0.4 & 12.0 \\
  & Two  & 83.8 & 91.7 & 71.2 & 88.8 & 82.3 & 75.3 & 94.4 & 62.0 & 72.9 & 67.4 & 80.5 & 13.0 & 0.0 & 1.1 & 11.8 \\
  & Four & 84.1 & 92.9 & 82.1 & 87.8 & 82.5 & 72.5 & 95.1 & 61.6 & 77.5 & 68.6 & 81.0 & 15.5 & 0.2 & 1.4 & 8.4 \\
\midrule
\multirow{3}{*}{Aya-8B} 
  & Zero & 77.1 & 85.8 & 62.4 & 83.8 & 77.9 & 73.0 & 78.0 & 62.6 & 57.4 & 63.6 & 80.5 & 0.1 & 0.0 & 0.0 & 6.3 \\
  & Two  & 77.4 & 88.6 & 69.6 & 88.0 & 77.8 & 73.4 & 79.5 & 59.7 & 64.4 & 67.9 & 81.3 & 11.5 & 0.0 & 1.1 & 7.2 \\
  & Four & 76.8 & 88.4 & 72.5 & 89.3 & 78.5 & 75.3 & 81.6 & 63.4 & 68.8 & 68.8 & 81.8 & 11.8 & 0.3 & 1.2 & 5.0 \\
\midrule
\rowcolor[gray]{0.9} \multicolumn{17}{l}{\textbf{14-32B Models}} \\
\multirow{3}{*}{Qwen2.5-14B} 
  & Zero & 86.6 & 92.0 & 75.5 & 91.4 & 85.8 & 89.6 & 94.0 & 76.4 & 71.2 & 63.6 & 79.9 & 0.1 & 0.1 & 1.0 & 19.5 \\
  & Two  & 85.6 & 93.4 & 78.9 & 91.3 & 85.8 & 89.1 & 95.9 & 71.2 & 78.6 & 69.0 & 80.8 & 18.3 & 0.3 & 2.0 & 18.1 \\
  & Four & 86.9 & 94.5 & 82.4 & 92.5 & 87.3 & 87.2 & 96.4 & 74.6 & 81.9 & 68.7 & 81.3 & 19.1 & 0.5 & 3.8 & 16.8 \\
\midrule
\multirow{3}{*}{Mistral-22B} 
  & Zero & 83.9 & 90.4 & 69.6 & 88.6 & 84.4 & 86.3 & 86.9 & 69.3 & 64.0 & 54.3 & 80.1 & 0.1 & 0.0 & 0.8 & 15.8 \\
  & Two  & 87.0 & 94.5 & 74.4 & 89.3 & 86.4 & 88.6 & 94.2 & 67.0 & 78.8 & 68.9 & 81.0 & 18.3 & 0.2 & 3.0 & 13.5 \\
  & Four & 86.9 & 95.2 & 79.7 & 89.8 & 87.2 & 89.0 & 95.6 & 69.1 & 80.6 & 69.3 & 81.5 & 0.7 & 0.2 & 2.5 & 5.1 \\
\midrule
\multirow{3}{*}{Qwen2.5-32B} 
  & Zero & 88.0 & 90.6 & 81.9 & 91.2 & 87.2 & 89.7 & 95.5 & 76.8 & 72.4 & 68.4 & 80.0 & 1.6 & 0.1 & 1.5 & 20.3 \\
  & Two  & 88.9 & 94.7 & 84.8 & 91.2 & 88.7 & 88.2 & 96.6 & 76.0 & 81.8 & 69.2 & 80.9 & 18.9 & 0.3 & 2.9 & 22.6 \\
  & Four & 88.9 & 94.7 & 85.3 & 91.8 & 89.3 & 87.9 & 97.0 & 79.0 & 84.1 & 68.9 & 81.4 & 19.4 & 0.7 & 2.8 & 19.1 \\
\midrule
\multirow{3}{*}{QwQ-32B} 
  & Zero & 82.2 & 89.6 & 77.1 & 88.9 & 81.5 & 84.0 & 92.5 & 70.8 & 60.6 & 63.4 & 79.4 & 1.1 & 0.2 & 1.0 & 18.0 \\
  & Two  & 88.1 & 94.4 & 82.9 & 89.7 & 87.9 & 87.8 & 95.9 & 71.8 & 82.0 & 58.7 & 80.3 & 14.9 & 0.1 & 1.9 & 19.4 \\
  & Four & 88.0 & 95.1 & 86.1 & 91.0 & 88.8 & 86.9 & 96.4 & 75.4 & 84.0 & 68.9 & 80.8 & 4.9 & 0.4 & 4.0 & 16.7 \\
\midrule
\multirow{3}{*}{Aya-32B} 
  & Zero & 81.2 & 90.5 & 61.6 & 89.7 & 82.3 & 85.5 & 83.1 & 70.3 & 66.0 & 68.8 & 79.3 & 0.1 & 0.1 & 0.5 & 19.3 \\
  & Two  & 85.4 & 94.9 & 74.4 & 91.1 & 85.0 & 80.4 & 92.3 & 65.5 & 75.0 & 70.7 & 80.2 & 17.5 & 0.1 & 4.6 & 14.4 \\
  & Four & 85.5 & 95.1 & 68.5 & 90.2 & 85.8 & 78.2 & 94.3 & 68.8 & 78.6 & 71.6 & 80.7 & 18.8 & 0.4 & 6.1 & 14.5 \\
\midrule
\rowcolor[gray]{0.9} \multicolumn{17}{l}{\textbf{70-72B Models}} \\
\multirow{3}{*}{llama3.3-70B} 
  & Zero & 88.0 & 94.1 & 76.8 & 91.8 & 90.0 & 91.9 & 92.9 & 76.8 & 70.9 & 64.2 & 80.0 & 0.1 & 0.0 & 0.7 & 20.2 \\
  & Two  & 90.1 & 96.7 & 83.2 & 93.1 & 90.1 & 85.3 & 96.4 & 74.7 & 79.8 & 71.7 & 81.0 & 16.5 & 0.3 & 6.1 & 16.1 \\
  & Four & 90.4 & 97.0 & 81.3 & 93.4 & 90.9 & 84.7 & 96.8 & 77.3 & 82.8 & 70.5 & 81.5 & 19.0 & 0.6 & 7.9 & 14.8 \\
\midrule
\multirow{3}{*}{Qwen2.5-72B} 
  & Zero & 89.1 & 92.6 & 84.3 & 92.6 & 89.4 & 92.1 & 93.4 & 77.5 & 75.6 & 68.4 & 79.9 & 0.1 & 0.0 & 1.0 & 21.6 \\
  & Two  & 90.5 & 95.0 & 85.9 & 93.2 & 90.8 & 88.4 & 96.5 & 73.2 & 82.5 & 69.4 & 80.9 & 15.1 & 0.3 & 1.2 & 20.7 \\
  & Four & 90.6 & 95.0 & 85.9 & 93.8 & 91.3 & 87.7 & 97.2 & 73.6 & 84.2 & 69.2 & 81.4 & 46.5 & 0.6 & 3.0 & 21.0 \\
\bottomrule
\end{tabular}
}
\caption{Performance of LLMs under zero-, two- and four-shot settings.}
\label{tab:full_few_shot_results}
\end{table*}

\begin{table*}
\small
\centering
\renewcommand{\arraystretch}{0.9}%
\setlength{\tabcolsep}{3.5pt}
\begin{tabular}{l | *{5}{r} | *{5}{r} | *{5}{r}}
\toprule
\textbf{Model}
& \multicolumn{5}{c|}{\textbf{Understanding} \scriptsize{(Acc.)}} 
& \multicolumn{5}{c|}{\textbf{Reasoning} \scriptsize{(Acc.)}} 
& \multicolumn{5}{c}{\textbf{Learning} \scriptsize{(BERTScore, F1)}} \\
\cmidrule(lr){2-6} \cmidrule(lr){7-11} \cmidrule(lr){12-16}
 & U1 & U2 & U3 & U4 & U5 & R1 & R2 & R3 & R4 & R5 & L1 & L2 & L3 & L4 & L5 \\
\midrule
\cellcolor[gray]{0.9} 3-4B &   &   &   &   &   &   &   &   &   &   &   &   &   &      &       \\
Qwen2.5-3B        & 74.3 & 82.0 & 76.3 & 82.0 & 74.0 & 77.5 & 71.2 & 62.0 & 59.0 & 54.0 & 85.4 & 0.4 & 0.2 & 0.3 & 9.0 \\
Phi4-4B           & 74.0 & 86.5 & 72.0 & 83.5 & 75.0 & 76.5 & 77.0 & 60.0 & 56.0 & 41.0 & 88.2 & 0.4 & 0.2 & 0.1 & 0.2 \\
\midrule
\cellcolor[gray]{0.9}7-8B  &   &   &   &   &   &   &   &   &   &   &   &   &   &      &       \\
Qwen2.5-7B \good  & 80.0 & 87.5 & 74.5 & 87.0 & 80.5 & 84.5 & 85.0 & 71.0 & 63.5 & 54.0 & 85.4 & 1.2 & 0.4 & 0.6 & 20.8 \\
Ollm-wiki-7B      & 71.0 & 81.5 & 64.5 & 78.5 & 73.5 & 62.0 & 80.0 & 50.5 & 54.5 & 47.5 & 84.5 & 0.4 & 0.2 & 0.3 & 0.2 \\
Ollm-arxiv-7B     & 70.8 & 81.3 & 64.0 & 78.3 & 73.5 & 61.0 & 79.5 & 50.0 & 53.5 & 46.7 & 84.6 & 0.4 & 0.2 & 0.1 & 11.0 \\
LLaMA3.1-8B       & 76.8 & 84.5 & 72.0 & 85.5 & 78.5 & 76.8 & 81.0 & 69.5 & 59.5 & 55.0 & 85.0 & 0.4 & 0.2 & 0.2 & 19.5 \\
Ministral-8B      & 76.0 & 85.8 & 60.0 & 81.0 & 76.5 & 78.5 & 85.5 & 57.5 & 59.5 & 42.0 & \textbf{88.4} & 0.4 & 0.2 & 0.2 & 20.8 \\
Internlm3-8B      & 80.0 & 87.5 & 69.5 & 85.5 & 79.5 & 85.5 & 87.5 & 71.0 & 64.0 & 50.0 & 85.3 & 0.5 & 0.2 & 0.7 & 16.0 \\
Aya-8B            & 74.0 & 82.8 & 60.0 & 80.8 & 75.0 & 70.0 & 75.0 & 60.0 & 55.0 & 50.0 & 86.1 & 0.3 & 0.1 & 0.1 & 8.5 \\
\midrule
\cellcolor[gray]{0.9}14-32B&   &   &   &   &   &   &   &   &   &   &   &   &   &      &       \\
Qwen2.5-14B       & 84.0 & 89.0 & 73.0 & 88.5 & 82.5 & 86.5 & 91.0 & 74.0 & 68.5 & 50.5 & 85.5 & 0.4 & 0.2 & 1.4 & 24.5 \\
Mistral-22B       & 81.0 & 87.5 & 67.0 & 86.0 & 81.5 & 83.5 & 84.0 & 66.5 & 61.0 & 43.0 & 85.7 & 0.4 & 0.2 & 1.2 & 20.5 \\
Qwen2.5-32B \good & 85.0 & 87.5 & 79.0 & 88.5 & 84.0 & 86.8 & \textbf{93.0} & 74.0 & 69.5 & 54.5 & 85.6 & \textbf{2.8} & 0.3 & 2.4 & 23.5 \\
QwQ-32B           & 79.5 & 86.0 & 74.5 & 86.0 & 78.5 & 81.5 & 89.5 & 68.0 & 58.0 & 49.0 & 85.0 & 2.0 & \textbf{0.4} & 1.6 & 22.0 \\
Aya-32B           & 78.5 & 87.5 & 59.0 & 86.5 & 79.0 & 82.5 & 80.0 & 67.5 & 63.0 & \textbf{54.5} & 84.9 & 0.4 & 0.2 & 0.9 & 23.8 \\
\midrule
\cellcolor[gray]{0.9}70-72B&   &   &   &   &   &   &   &   &   &   &   &   &   &      &       \\
LLaMA3.3-70B      & 86.0 & \textbf{92.0} & 75.0 & 90.0 & \textbf{88.5} & 90.0 & 91.0 & 75.0 & 69.0 & 50.5 & 85.6 & 0.3 & 0.2 & 1.2 & 23.0 \\
Qwen2.5-72B \best & \textbf{87.5} & 90.5 & \textbf{82.5} & \textbf{90.5} & 87.5 & \textbf{90.5} & 91.5 & \textbf{76.0} & \textbf{73.0} & 54.5 & 85.5 & 0.3 & 0.2 & 1.8 & \textbf{24.0} \\
\bottomrule
\end{tabular}
\caption{Performance of LLMs under zero-shot chain-of-thought prompting results}
\label{tab:zero-shot-cot-performance}
\end{table*}

\section{License}
Because \textsc{OntoURL} uses open source data, its license is Creative Commons Attribution 4.0 International (CC BY 4.0)—you’re free to share and adapt the dataset provided that you give appropriate credit to the original source.





  

\end{document}